\documentclass[runningheads]{llncs}
\usepackage{graphicx}
\usepackage{amssymb}
\usepackage{amsmath}
\usepackage{amsfonts}
\usepackage{pifont}
\usepackage{multirow, ltablex, xspace}
\usepackage[table]{xcolor}
\usepackage{stmaryrd}
\usepackage{hyperref}

\newcommand{\mymodel}{HADA}

\begin{document}
%
%
%
\title{HADA: A Graph-based Amalgamation Framework in Image-text Retrieval}
\titlerunning{HADA in Image-text Retrieval}
\author{Manh-Duy Nguyen\inst{1}\orcidID{0000-0001-6878-7039} \and
Binh T. Nguyen\inst{2,3,4} \and
Cathal Gurrin\inst{1}\orcidID{0000-0003-2903-3968}}
\authorrunning{Manh-Duy et al.}
%
\institute{School of Computing, Dublin City University, Ireland \and VNU-HCM, University of Science, Vietnam \and Vietnam National University Ho Chi Minh City, Vietnam \and AISIA Lab, Vietnam}
\maketitle              
\begin{abstract}
Many models have been proposed for vision and language tasks, especially the image-text retrieval task. 
All state-of-the-art (SOTA) models in this challenge contained hundreds of millions of parameters.
They also were pretrained on a large external dataset that has been proven to make a big improvement in overall performance. 
It is not easy to propose a new model with a novel architecture and intensively train it on a massive dataset with many GPUs to surpass many SOTA models, which are already available to use on the Internet.
In this paper, we proposed a compact graph-based framework, named \mymodel, which can combine pretrained models to produce a better result, rather than building from scratch.
First, we created a graph structure in which the nodes were the features extracted from the pretrained models and the edges connecting them. 
The graph structure was employed to capture and fuse the information from every pretrained model with each other.
Then a graph neural network was applied to update the connection between the nodes to get the representative embedding vector for an image and text. 
Finally, we used the cosine similarity to match images with their relevant texts and vice versa to ensure a low inference time. 
Our experiments showed that, although \mymodel\ contained a tiny number of trainable parameters, it could increase baseline performance by more than $3.6\%$ in terms of evaluation metrics in the Flickr30k dataset.
Additionally, the proposed model did not train on any external dataset and did not require many GPUs but only 1 to train due to its small number of parameters. The source code is available at \url{https://github.com/m2man/HADA}.

\end{abstract}


\section{Introduction}
Image-text retrieval is one of the most popular challenges in vision and language tasks, with many state-of-the-art models (SOTA) recently introduced \cite{li2020oscar,chen2020uniter,li2021albef,radford2021clip,gan2020villa,miech2021fastandslow,li2022blip}.
This challenge includes 2 subtasks, which are image-to-text retrieval and text-to-image retrieval. 
The former subtask is defined as an image query is given to retrieve relevant texts in a multimodal dataset, while the latter is vice versa.

\begin{figure*}[ht!]
  \centering
  \includegraphics[width=0.99\linewidth]{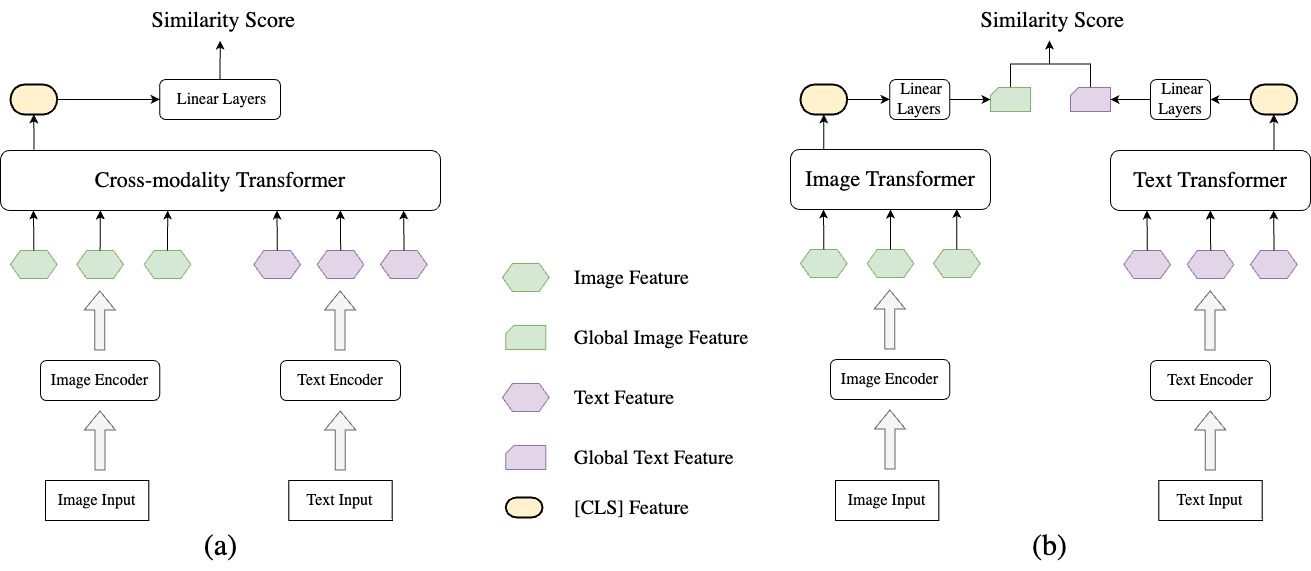}
  \caption{Two most popular pipelines of the SOTA for image-text retrieval challenge. (a) A cross-modality transformer network is applied to measure the similarity between an image and a text based on their features. (b) Each modality used their own transformer network to get its global embedding.}
  \label{fig:intro}
\end{figure*}

Most of the SOTA models in this research field shared 2 things in common: (1) they were built on transformer-based cross-modality attention architectures \cite{chen2020uniter,li2020oscar} and (2) they were pretrained on the large-scale multimodal data crawled from the Internet \cite{radford2021clip,li2021albef,li2020oscar,li2022blip,jia2021scaling}. 
However, these things have their own disadvantages. 
The attention structure between 2 modalities could achieve an accurate result, but it cost a large amount of inference time due to the massive computation. For instance, UNITER \cite{chen2020uniter} contained roughly 303 millions parameters and it took a decent amount of time to perform the retrieval in real-time \cite{sun2021lightningdot}. 
Many recent work has resolved this model-related problem by introducing joint-encoding learning methods.
They can learn visual and semantic information from both modalities without using any cross-attention modules, which can be applied later to rerank the initial result \cite{li2021albef,miech2021fastandslow,sun2021lightningdot}. 
Figure \ref{fig:intro} illustrated the architecture of these pipelines. 
Regarding the data perspective, the large collected data usually come with noisy annotation, and hence could be harmful to the models that are trained on it. Several techniques have been proposed to mitigate this issue \cite{li2020oscar,li2021albef,li2022blip}. However, training on the massive dataset still creates a burden on the computation facility, such as the number of GPUs, which are required to train the model successfully and efficiently \cite{radford2021clip}.

It has motivated us to answer the question: \textit{Can we combine many SOTA models, which are currently available to use, to get a better unified model without intensively training with many GPUs?} 
In this paper, we introduced a grap\underline{\textbf{h}}-b\underline{\textbf{a}}se\underline{\textbf{d}} \underline{\textbf{a}}malgamation framework, called \textbf{\mymodel}, which formed a graph-based structure to fuse the features produced by other pretrained models. 
We did not use any time-consuming cross-modality attention network to ensure fast retrieval speed. 
A graph neural network was employed to extract visual and textual embedded vectors from fused graph-based structures of images and texts, where we can measure their cosine similarity. 
To the best of our knowledge, the graph structure has been widely applied in the image-text retrieval challenge \cite{nguyen2021deep,diao2021similarity,nguyen2021graph,wang2020cross,liu2020graph}. Nevertheless, it was utilized to capture the interaction between objects or align local and global information within images. \mymodel\ is the first approach that applies this data structure to combine SOTA pretrained models by fusing their features in each modality.
We trained \mymodel\ only on the Flickr30k dataset without using any large-scale datasets. We applied Momentum Distillation technique \cite{li2021albef}, which was shown that can not only mitigate the harmful effect of noise annotation but also improve the accuracy on a clean dataset.
Our experiments showed that \mymodel, with the tiny extra number of training parameters, could improve total recalls by $3.64\%$ compared to the input SOTA without training with millions of additional image-text pairs as other models.
This is the most crucial part since it is not easy to possess multiple GPUs to use, especially for small and medium businesses or start-up companies. Therefore, we believe that \mymodel\ can be applied not only in the academic field, but also in industry.

Our main contribution can be summarised as follow: (1) We introduced \mymodel, a compact pipeline that can combine 2 or many SOTA pretrained models to address the image-text retrieval challenge. (2) We proposed a way to fuse the information between input pretrained models by using graph structures. (3) We evaluated the performance of \mymodel\ on the well-known Flickr30k dataset \cite{flickr30k} and MSCOCO dataset \cite{2014mscoco} without using any other large-scale dataset but still improved the accuracy compared to the baseline input models.

\section{Related Work}
A typical vision-and-language model, including image-text retrieval task, was built with the usage of transformer-based encoders. In specific, OSCAR \cite{li2020oscar}, UNITER \cite{chen2020uniter}, and VILLA \cite{gan2020villa} firstly employed Faster-RCNN \cite{ren2015fasterrcnn} and BERT \cite{devlin2018bert} to extract visual and text features from images and texts. These features were then fed into a cross-modality transformer block to learn the contextualized embedding that captured the relations between regional features from images and word pieces from texts. An additional fully connected layer was used to classify whether the images and texts were relevant to each other or not, based on the embedding vectors. Although achieving superior results, these approaches had a drawback of being applied to real-time use cases. It required a huge amount of time to perform the retrieval online, since the models have to process the intensive cross-attention transformer architecture many times for every single query \cite{sun2021lightningdot}.

Recently, there have been some works proposing an approach to resolve that problem by utilising 2 distinct encoders for images and text. The data from each modality now can be embedded offline and hence improve the retrieval speed \cite{sun2021lightningdot,li2021albef,li2022blip,miech2021fastandslow,jia2021scaling,radford2021clip}. In terms of architecture, all approaches used the similar BERT-based encoder for semantic data but different image encoders. While LightningDOT \cite{sun2021lightningdot} encoded images with detected objects extracted by the Faster-RCNN model, FastnSlow \cite{miech2021fastandslow} applied the conventional Resnet network to embed images. On the other side, ALBEF \cite{li2021albef} and BLIP \cite{li2022blip} employed the Vision Transformer backbone \cite{dosovitskiy2020vist} to get the visual features corresponding to their patches. Because these SOTA did not use the cross-attention structure, which was a critical point to achieve high accuracy, they applied different strategies to increase performance. Specifically, pretraining a model on a large dataset can significantly improve the result \cite{li2021albef,li2020oscar,jia2021scaling}. For instance, CLIP \cite{radford2021clip} and ALIGN \cite{jia2021scaling} were pretrained on 400 millions and 1.8 billions image-text pairs, respectively. Another way was that they ran another cross-modality image-text retrieval model to rerank the initial output and get a more accurate result \cite{li2021albef,sun2021lightningdot}.

Regarding to graph structures, SGM \cite{wang2020cross} introduced a visual graph encoder and a textual graph encoder to capture the interaction between objects appearing in images and between the entities in text. LGSGM \cite{nguyen2021deep} proposed a graph embedding network on top of SGM to learn both local and global information about the graphs. Similarly, GSMN \cite{liu2020graph} presented a novel technique to assess the correspondence of nodes and edges of graphs extracted from images and texts separately. SGRAF \cite{diao2021similarity} build a reasoning and filtration graph network to refine and remove irrelevant interactions between objects in both modalities.

Although there are many SOTAs with different approaches for image-text retrieval problems, there is no work that tries combining these models but introducing a new architecture and pretrain on a massive dataset instead. Training an entire new model from scratch on the dataset is not an easy challenge since it will create a burden on the computation facilities such as GPUs. In this paper, we introduced a simple method which combined the features extracted from the pretrained SOTA by applying graph structures. Unlike other methods that also used this data structure, we employed graphs to fuse the information between the input features, which was then fed into a conventional graph neural network to obtain the embedding for each modality. Our \mymodel\ consisted of a small number of trainable parameters, hence can be easily trained on a small dataset but still obtained higher results compared to the input models.

\section{Methodology}
\begin{figure*}[ht!]
  \centering
  \includegraphics[width=0.99\linewidth]{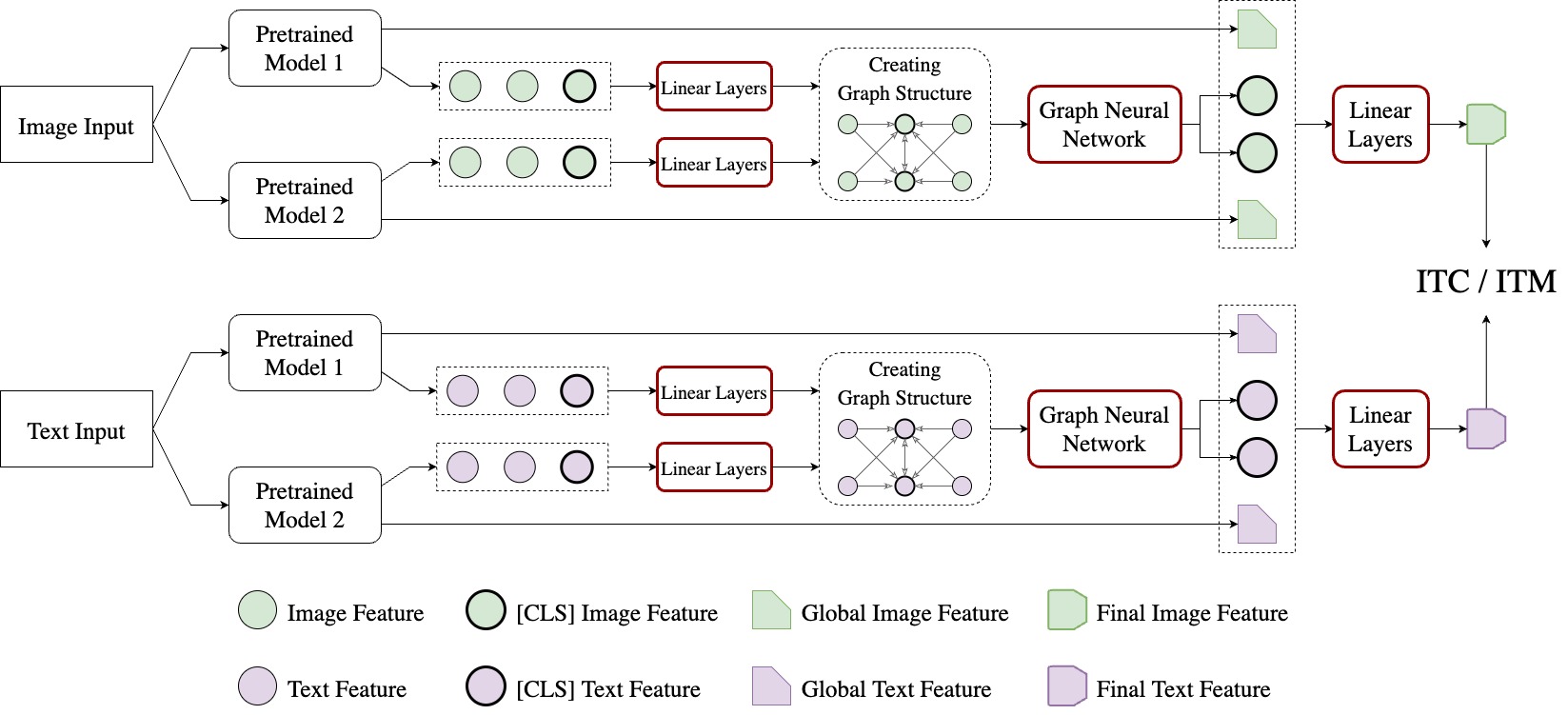}
  \caption{The pipeline of the proposed \mymodel. The red borders indicated trainable components. The ITM and ITC infered the training tasks which will be discussed later.}
  \label{fig:method}
\end{figure*}

This section will describe how our \mymodel\ addressed the retrieval challenge
by combining any available pretrained models. 
Figure \ref{fig:method} depicted the workflow of \mymodel. We started with only 2 models $(N_{models}=2)$ as illustrated in Figure \ref{fig:method} for simplicity. Nevertheless, \mymodel\ can be extended with a larger $N_{models}$.
\mymodel\ began with using some pretrained models to extract the features from each modality. We then built a graph structure to connect the extracted features together, which were fed into a graph neural network (GNN) later to update them. 
The outputs of the GNN were concatenated with the original global features produced by the pretrained models. 
Finally, simple linear layers were employed at the end to get the final representation embedding features for images and texts, which can be used to measure the similarity to perform the retrieval.
For evaluation, we could extract our representation features offline to guarantee the high speed inference time.

\subsection{Revisit State-of-the-art Models}
We only used the pretrained models without using the cross-modality transformer structure to extract the features as depicted in Figure \ref{fig:intro} to reduce the number of computations and ensure the high speed inference time. Basically, they used a unimodal encoder to get the features of an image or a text followed by a transformer network to embed them and obtain the \textit{[CLS]} embedding. This \textit{[CLS]} token was updated by 1 or many fully connected layers to become a representative global feature that can be compared with that of the remaining modality to get the similarity score. 

\mymodel\ began with the output of the transformer layer from the pretrained models. 
In detail, for an input image \textit{\textbf{I}}, we obtained the sequence of patch tokens from each model $i$ denoted as  $\mathbf{v^{(i)}} = \{v^{(i)}_{cls}, v^{(i)}_1, v^{(i)}_2, ... ,v^{(i)}_{N_i}\}$, where $v^{(i)}_j \in \mathbb{R}^{d^{(i)}_{v}}$ and $N_i$ was the length of the sequence. 
This length depended on the architecture of the image encoder network employed in the pretrained model. For example, it could be the number of patches if the image encoder was a Vision Transformer (ViT) network \cite{dosovitskiy2020vist}, or the number of detected objects or regions of interest if the encoder was a Faster-RCNN model \cite{ren2015fasterrcnn}. 
Additionally, we also extracted the global visual representation feature $h^{(i)}_v \in \mathbb{R}^{d^{(i)}_h}$ from $v^{(i)}_{cls}$ as illustrated in Figure \ref{fig:intro}. 
Regarding the semantic modality, we used the same process as that of the visual modality. 
Specifically, we extracted the sequence of patch tokens $\mathbf{w^{(i)}} = \{w^{(i)}_{cls}, w^{(i)}_1, w^{(i)}_2, ... ,w^{(i)}_{L}\}$  where $w^{(i)}_j \in \mathbb{R}^{d^{(i)}_{w}}$ and $L$ was the length of the text, and the global textual representation embedding $h^{(i)}_w \in \mathbb{R}^{d^{(i)}_h}$ for an input text \textit{\textbf{T}} using the pretrained model $i$. 
The input model $i$ matched a pair of an image \textit{\textbf{I}} and a text \textit{\textbf{T}} by calculating the dot product $\langle h^{(i)}_v , h^{(i)}_w\rangle$ of their global features. 
However, \mymodel\ not only used the global embedding but also the intermediate transformer tokens to make the prediction. 
We used our learned \textit{[CLS]} tokens to improve the global features. 
In contrast, using the original global features could ensure high performance of the pretrained models and mitigate the effect of unhelpful tokens.

\subsection{Create Graph Structure}
Each pretrained model \textit{i} produced different \textit{[CLS]} features $v^{(i)}_{cls}$ and $w^{(i)}_{cls}$ for an image and text, respectively. 
Since our purpose was to combine the models, we needed to fuse these \textit{[CLS]} tokens to obtain the unified ones for each modality separately. 
In each modality, for example, the visual modality, \mymodel\ not only updated $v^{(i)}_{cls}$ based on $\mathbf{v^{(i)}}$ solely but also on those of the remaining pretrained models $\{\mathbf{v^{(j)}} \ | \ j \neq i\}$. 
Because these $\mathbf{v}$ came from different models, their dimensions could be not similar to each other. 
Therefore, we applied a list of linear layers $f^{(i)}_v:\mathbb{R}^{d^{(i)}_{v}}\rightarrow\mathbb{R}^{d_p}$ to map them in the same dimensional space: $$\mathbf{p^{(i)}} = \{f^{(i)}_v(x) | x \in \mathbf{v^{(i)}}\} = \{p^{(i)}_{cls}, p^{(i)}_1, p^{(i)}_2, ... ,p^{(i)}_{N_i}\}$$
We performed a similar process for the textual modality to obtain: $$\mathbf{s^{(i)}} = \{f^{(i)}_w(x) | x \in \mathbf{w^{(i)}}\} = \{s^{(i)}_{cls}, s^{(i)}_1, s^{(i)}_2, ... ,s^{(i)}_L\}, \textnormal{where} \ f^{(i)}_w:\mathbb{R}^{d^{(i)}_{w}}\rightarrow\mathbb{R}^{d_s}$$

We then used graph structures $\mathcal{G}p = \{\mathcal{V}_p,\mathcal{E}_p\}$ and $\mathcal{G}_s = \{\mathcal{V}_s,\mathcal{E}_s\}$ to connect these mapped features together, where $\mathcal{V}$ and $\mathcal{E}$ denoted the list of nodes and edges in the graph $\mathcal{G}$ accordingly. 
In our \mymodel, nodes indicated the mapped features. Specifically, $\mathcal{V}_p = \{\mathbf{p^{(i)}}\}$ and $\mathcal{V}_s = \{\mathbf{s^{(i)}}\}$ for all $i \in [1,N_{models}]$.
Regarding edges, we symbolized $e_{a \rightarrow b}$ as a directed edge from node $a$ to node $b$ in the graph, thus the set of edges of the visual graph $\mathcal{E}_p$ and the textual graph $\mathcal{E}_s$ were:
$$\mathcal{E}_p=\{e_{{x} \rightarrow p^{(j)}_{cls}} \ | \ x \in \mathbf{p^{(i)}} \ \textnormal{and} \ i,j \in [1,N_{models}]\}$$
$$\mathcal{E}_s=\{e_{{x} \rightarrow s^{(j)}_{cls}} \ | \ x \in \mathbf{s^{(i)}} \ \textnormal{and} \ i,j \in [1,N_{models}]\}$$
To be more detailed, we created directed edges that went from every patch features to the \textit{[CLS]} feature, including from the \textit{[CLS]} itself, for all pretrained models but not in the reversed direction, as shown in Figure \ref{fig:method}. 
The reason was that \textit{[CLS]} was originally introduced as a representation of all input data, so it would summarize all patch tokens \cite{dosovitskiy2020vist,chen2021crossvit,devlin2018bert}.
Therefore, it would be the node that received information from other nodes in the graph. 
This connection structure ensured that \mymodel\ can update the \textit{[CLS]} tokens based on the patch tokens from all pretrained models in a fine-grained manner.

\subsection{Graph Neural Network}
Graph neural networks (GNN) have witnessed an increase in its popularity over the past few years, with many GNN structures having been introduced recently \cite{kipf2016semi,defferrard2016convolutional,velivckovic2017graph,hamilton2017inductive,shi2020masked,brody2021attentive}. 
\mymodel\ applied the modified Graph Attention Network (GATv2), which was recommended to be used as a baseline whenever employing GNN \cite{brody2021attentive}, to fuse the patch features from different pretrained models together to get the unified \textit{[CLS]} features.
Let $\mathcal{N}_k = \{x \in \mathcal{V} \ | \ e_{{x} \rightarrow k} \in \mathcal{E}\}$ be the set of neighbor nodes from which there was an edge connecting to node $k$ in the graph $\mathcal{G}$. 
GATv2 used a scoring function $se$ to weight every edge indicating the importance of the neighbor nodes $x$ in $\mathcal{N}_k$ before updating the node $k \in \mathbb{R}^d$:
$$se(e_{{x} \rightarrow k}) = \mathbf{A}^{\top} \textnormal{LeakyRELU}(\mathbf{W}_1x +\mathbf{W}_2k])$$
where $\mathbf{A}\in\mathbb{R}^{d'}$, $\mathbf{W}_1 \in \mathbb{R}^{d' \times d}$, and $\mathbf{W}_2 \in \mathbb{R}^{d' \times d}$ were learnable parameters. These weights were then normalized across all neighbor nodes in $\mathcal{N}_k$ by using a softmax function to get the attention scores:
$$\alpha_{e_{{x} \rightarrow k}} = \frac{\textnormal{exp}(se(e_{{x} \rightarrow k}))}{\sum_{y\in\mathcal{N}_k}^{}\textnormal{exp}(se(e_{{y} \rightarrow k}))}$$
The updated node $k' \in \mathbb{R}^{d'}$ was then calculated based on its neighbors in $\mathcal{N}_k$, including $k$ if we add an edge connect it to itself:
$$k' = \sigma(\sum_{x\in\mathcal{N}_k}^{}\alpha_{e_{{x} \rightarrow k}} \cdot \mathbf{W}_1x)$$
where $\sigma$ was an nonlinearity activate function. Furthermore, this GATv2 network could be enlarged by applying a multi-head attention structure and improved performance \cite{velivckovic2017graph}. 
The output now was a concatenation of each head output, which was similar to Transformer architecture \cite{vaswani2017attention}. An extra linear layer was used at the end to convert these concatenated nodes to the desired dimensions.

We used distinct GATv2 structures with $H$ attention heads for each modality in this stage, as illustrated in Figure \ref{fig:method}. 
\mymodel\ took the input graphs $\mathcal{G}_p$ and $\mathcal{G}_s$ with nodes $\mathcal{V}_p$ and $\mathcal{V}_s$ in the vector space of $d_p$ and $d_s$ dimensions and updated them to $\mathcal{V'}_p=\{\mathbf{p'^{(i)}}\}$ and $\mathcal{V'}_s=\{\mathbf{s'^{(i)}}\}$ with dimensions of $d'_p$ and $d'_s$.
We then concatenated the updated \textit{[CLS]} nodes $p'_{cls}$ and $s'_{cls}$ from all pretrained models with their corresponding original global embedding $h_v$ and $h_w$. Finally, we fed them into a list of linear layers to get our normalized global representation $h_p \in \mathbb{R}^{d_h}$ and $h_s \in \mathbb{R}^{d_h}$.

\subsection{Training Tasks}
\subsubsection{Image-Text Contrastive Learning.} \mymodel\ encoded the input image \textit{\textbf{I}} and text \textit{\textbf{T}} to $h_p$ and $h_s$, accordingly. We used a similarity function that was a dot product $S(\textnormal{\textit{\textbf{I}}},\textnormal{\textit{\textbf{T}}}) = \langle h_p, h_s \rangle = h^{\top}_p h_s$ to ensure that a pair of relevant image-text (positive pair) would have a higher similar representation compared to irrelevant pairs (negative pairs). 
The contrastive loss for image-to-text (i2t) retrieval and text-to-image (t2i) retrieval for the mini-batch of $M$ relevant pairs $(\textnormal{\textit{\textbf{I}}}_m, \textnormal{\textit{\textbf{T}}}_m)$ were:
$$\mathcal{L}_{i2t}(\textnormal{\textit{\textbf{I}}}_m) = -\textnormal{log}\frac{\textnormal{exp}(S(\textnormal{\textit{\textbf{I}}}_m,\textnormal{\textit{\textbf{T}}}_m)/\tau)}{\sum^{M}_{i=1} \textnormal{exp}(S(\textnormal{\textit{\textbf{I}}}_m,\textnormal{\textit{\textbf{T}}}_i)/\tau)}$$
$$\mathcal{L}_{t2i}(\textnormal{\textit{\textbf{T}}}_m) = -\textnormal{log}\frac{\textnormal{exp}(S(\textnormal{\textit{\textbf{T}}}_m,\textnormal{\textit{\textbf{I}}}_m)/\tau)}{\sum^{M}_{i=1} \textnormal{exp}(S(\textnormal{\textit{\textbf{T}}}_m,\textnormal{\textit{\textbf{I}}}_i)/\tau)}$$
where $\tau$ was a temperature parameter that could be learned during training. 
This contrastive learning has been used in many vision-and-language models and has been proven to be effective \cite{li2021albef,sun2021lightningdot,li2022blip,radford2021clip}.
In our experiment, we trained \mymodel\ with the loss that optimized both subtasks:
$$\mathcal{L}_{ITC} = \frac{1}{M}\sum^{M}_{m=1}(\mathcal{L}_{i2t}(\textnormal{\textit{\textbf{I}}}_m) + \mathcal{L}_{t2i}(\textnormal{\textit{\textbf{T}}}_m))$$

Inspired by ALBEF \cite{li2021albef}, we also applied momentum contrast (MoCo) \cite{he2020momentum} and their momentum distillation strategy for this unsupervised representation learning to cope with the problem of noisy information in the dataset and improve accuracy. 

\subsubsection{Image-Text Matching}
This objective was a binary classification task to distinguish irrelevant image-text pairs, but were similar representations. 
This task would ensure that they were different in fine-grained details. We implemented an additional disciminator layer $dc:\mathbb{R}^{4d_h} \rightarrow \mathbb{R}$ on top of the final embedding features $h_p$ and $h_s$ to classify whether the image \textit{\textbf{I}} and the text \textit{\textbf{T}} is a positive pair or not:
$$dc(h_p,h_s) = \textnormal{sigmoid}(\mathbf{W}^{\top}[h_p \Vert  h_s \Vert  \textnormal{abs}(h_p - h_s) \Vert  h_p \odot h_s])$$
where  $\mathbf{W} \in \mathbb{R}^{4d_h}$ was trainable parameters, $\Vert$ indicated the concatenation, $\textnormal{abs(.)}$ was the absolute value, and $\odot$ denoted elementwise multiplication. 
We used binary cross-entropy loss for this ordinary classification task:
$$\mathcal{L}_{itm}(\textnormal{\textit{\textbf{I}}},\textnormal{\textit{\textbf{T}}}) = y\textnormal{log}(dc(h_p,d_s)) + (1-y)\textnormal{log}(1-dc(h_p,d_s))$$
where $y$ was the one-hot vector representing the ground truth label of the pair.

For each positive pair in the minibatch of $M$ positive pairs, we sampled 1 hard negative text for the image and 1 hard negative image for the text.
These negative samples were chosen from the current mini-batch in which they were not relevant based on the ground-truth labels, but have the highest similarity dot product score. 
Therefore, the objective for this task was:
$$\mathcal{L}_{ITM} = \frac{1}{3M}\sum^{M}_{m=1}(\mathcal{L}_{itm}(\textnormal{\textit{\textbf{I}}}_m,\textnormal{\textit{\textbf{T}}}_m) + \mathcal{L}_{itm}(\textnormal{\textit{\textbf{I}}}_m,\textnormal{\textit{\textbf{T}}}'_m) + \mathcal{L}_{itm}(\textnormal{\textit{\textbf{I}}}'_m,\textnormal{\textit{\textbf{T}}}_m))$$
where $\textnormal{\textit{\textbf{T}}}'_m$ and $\textnormal{\textit{\textbf{I}}}'_m$ were the hard negative text and image samples in the mini-batch that were corresponding with the $\textnormal{\textit{\textbf{I}}}_m$ and $\textnormal{\textit{\textbf{T}}}_m$, respectively.
The final loss function in \mymodel\ was:
$$\mathcal{L} = \mathcal{L}_{ITC} + \mathcal{L}_{ITM}$$


\section{Experiment}
\subsection{Dataset and Evaluation Metrics}
We trained and evaluated \mymodel\ on 2 different common datasets in the image-text retrieval task which are Flickr30k \cite{flickr30k} and MSCOCO \cite{2014mscoco}.
Flickr30k dataset consists of 31K images collected on the Flickr website, while MSCOCO comprises 123K images.
Each image contains 5 relevant texts or captions that describe the image.
We used Karpathy's split \cite{karpathy2015deep}, which has been widely applied by all models in the image-text retrieval task, to split each dataset into train/evaluate/test on 29K/1K/1K and 113K/5K/5K images on Flickr30k and MSCOCO, respectively. 

The common evaluation metric in this task is the Recall at K ($R@K$) when many SOTAs used this metric \cite{li2021albef,sun2021lightningdot,li2022blip,radford2021clip,jia2021scaling,li2020oscar,chen2020uniter,gan2020villa}. 
This metric is defined as the proportion of the number of queries that we found the correct relevant output in the top K of the retrieved ranked list:
$$R@K = \frac{1}{N_q}\sum^{N_q}_{q=1}\textbf{1}(q, K)$$
where $N_q$ is the number of queries and $\textbf{1}(q,K)$ is a binary function returning 1 if the model find the correct answer of the query $q$ in the top $K$ of the retrieved output.
In particular, for the image-to-text subtask, $R@K$ is the percentage of the number of images where we found relevant texts in the top K of the output result.
In our experiment, we used R@1, R@5, R@10, and RSum which was the sum of them.

\subsection{Implementation Details} \label{writing:implement}
In our experiment, we combined 2 SOTA models that had available pretrained weights fine-tuned on the Flickr30k dataset: ALBEF\footnote{\url{https://github.com/salesforce/ALBEF}} and LightningDOT\footnote{\url{https://github.com/intersun/LightningDOT}}.
None of them used the cross-modality transformer structure when retrieved to ensure the fast inference speed\footnote{Indeed, these 2 models applied the cross-modality transformer network to rerank the initial result in the subsequent step. However, we did not focus on this stage.}. 
Although they used the same BERT architecture to encode a text, the former model employed the ViT network to encode an image, while the latter model applied the Faster-RCNN model.
We chose these 2 models because we wanted to combine different models with distinct embedding backbones to utilize the advantages of each of them.

Regarding ALBEF, their ViT network encoded an image to $577$ patch tokens including the \textit{[CLS]} one ($N_{ALB}=576$ and $d^{(ALB)}_v = 768$). This \textit{[CLS]} was projected to the lower dimension to obtain the global feature ($d^{(ALB)}_h=256$). 
Because LightningDOT encoded an image based on the detected objects produced by the Faster-RCNN model, its $N_{DOT}$ varied depending on the number of objects in the image.
The graph neural network, unlike other conventional CNN, can address this inconsistent number of inputs due to the flexible graph structure with nodes and edges.
Unlike ALBEF, the dimensions of image features and global features from LightningDOT were the same with $d^{(DOT)}_v = d^{(DOT)}_h = 768$. 
In terms of text encoder, the output of both models was similar since they used the same BERT network: $d^{(ALB)}_w = d^{(DOT)}_w = 768$. 
We projected these features to a latent space where $d_p = d_s = 512$, which were the average of their original dimensions. 
We used a 1-layer GATv2 network with $H = 4$ multi-head attentions to update the graph features while still keeping the input dimensions of $d'_p = d'_s = 512$. We also applied Dropout with $p = 0.7$ in linear layers and graph neural networks. In total, our \mymodel\ contained roughly 10M trainable parameters.

The input pretrained models were pretrained on several large external datasets. 
For example, ALBEF was pretrained on 14M images compared to only 29K images on Flickr30k that we used to train \mymodel. 
We used this advantage in our prediction instead of train \mymodel\ in millions of samples. 
We modified the similarity score to a weighted sum of our predictions and the original prediction of the input models.
Therefore, the weighted similarity score that we used was:
$$S(\textnormal{\textit{\textbf{I}}},\textnormal{\textit{\textbf{T}}}) = (1-\alpha) \langle h_p,h_s \rangle + \alpha \langle h^{(ALB)}_v,h^{(ALB)}_w \rangle$$
where $\alpha$ was a trainable parameter. We did not include the original result of the LightningDOT model, since its result was lower than ALBEF by a large margin and therefore could have a negative impact on overall performance\footnote{We tried including the LightningDOT in the weighted similarity score, but the result was lower than using only ALBEF.}. 

We trained \mymodel\ for 50 epochs (early stopping\footnote{In our experiment, it converged after roughly 20 epochs.} was implemented) using the batch size of 20 on 1 NVIDIA RTX3080Ti GPU. 
We used the AdamW \cite{loshchilov2017decoupled} optimizer with a weight decay of 0.02. 
The learning rate was set at $1e^{-4}$ and decayed to $5e^{-6}$ following cosine annealing \cite{loshchilov2016sgdr}. 
Similarly to ALBEF, we also applied RandAugment \cite{cubuk2020randaugment} for data augmentation. 
The initial temperature parameter was $0.07$ \cite{wu2018unsupervised} and we kept it in range of $[0.001, 0.5]$ during training. 
To mitigate the dominant effect of ALBEF global features on our weighted similarity score, we first trained \mymodel\ with $\alpha = 0$. 
After the model had converged, we continued to train, but initially set $\alpha = 0.5$ and kept it in the range of $[0.1, 0.9]$.

\subsection{Baselines}
We built 2 baselines that also integrated ALBEF and LightningDOT as an input to show the advantages of using graph structures to fuse these input models.

\subsubsection{Baseline B1.} We calculated the average of the original ranking results obtained from ALBEF and LightningDOT and considered them as the distance between images and text. This meant that the relevant pairs should be ranked at the top, whilst irrelevant pairs would have lower places. 

\subsubsection{Baseline B2.} Instead of using a graph structure to fuse the features extracted from the pretrained models, we only concatenated their global embedding and fed them into the last linear layers to obtain the unified features. We trained this baseline B2 following the same strategy as described in Section \ref{writing:implement} using the weighted similarity score.

\subsection{Comparison to Baseline}
Table \ref{tab:flickr30k} illustrated the evaluation metrics of the difference models in the Flickr30k dataset. Similarly to LightningDOT, our main target was to introduce an image-text retrieval model that did not implement a cross-modality transformer module to ensure that it can perform in real-time without any delay. Thus, we only reported the result from LightningDOT and ALBEF that did not use the time-consuming compartment to rerank in the subsequent step. If the model has a better initial result, it can have a better reranked result by using the cross-modality transformer later. We also added UNITER \cite{chen2020uniter} and VILLA \cite{gan2020villa}, which both used cross-modality transformer architecture to make the prediction, to the comparison.

\begin{table}[ht!]
  \centering
  \caption{Performance of models on Flickr30k Dataset. The symbol \ding{61} indicated the results were originally reported in their research, while others were from our re-implementation using their public pretrained checkpoints. The column $\vartriangle$R showed the difference compared to ALBEF.}
  \label{tab:flickr30k}
  \begin{tabular}{c|c c c c | c c c c|c|c}
    \hline
    \multirow{2}{*}{\textbf{Methods}} & \multicolumn{4}{c}{\textbf{Image-to-Text}} & \multicolumn{4}{c|}{\textbf{Text-to-Image}} & 
    \textbf{Total} & 
    \multirow{2}{*}{$\vartriangle$\textbf{R}} \\
    \cline{2-10}
    & R@1 & R@5 & R@10 & RSum & R@1 & R@5 & R@10 & RSum & RSum & \\
    \hline
    $\textnormal{UNITER}^{\textnormal{\ding{61}}}$ & 87.3 & 98 & 99.2 & 284.5 & 75.56 & 94.08 & 96.76 & 266.4 & 550.9 & \textcolor{red}{$\shortdownarrow$13.68} \\
    $\textnormal{VILLA}^{\textnormal{\ding{61}}}$ & 87.9 & 97.2 & 98.8 & 283.9 & 76.26 & 94.24 & 96.84 & 267.34 & 551.24 & \textcolor{red}{$\shortdownarrow$13.34}\\
    \hline
    LightningDOT & 83.6 & 96 & 98.2 & 277.8 & 69.2 & 90.72 & 94.54 & 254.46 & 532.26 & \textcolor{red}{$\shortdownarrow$32.32} \\
    $\textnormal{LightningDOT}^{\textnormal{\ding{61}}}$ & 83.9 & 97.2 & 98.6 & 279.7 & 69.9 & 91.1 & 95.2 & 256.2 & 535.9 & \textcolor{red}{$\shortdownarrow$28.68}\\
    ALBEF & 92.6 & 99.3 & 99.9 & 291.8 & 79.76 & 95.3 & 97.72 & 272.78 & 564.58 & 0 \\
    \hline
    B1 & 90.7 & 99 & 99.6 & 289.3 & 79.08 & 94.5 & 96.94 & 270.52 & 559.82 & \textcolor{red}{$\shortdownarrow$4.76}\\
    B2 & 91.4 & 99.5 & 99.7 & 290.6 & 79.64 & 95.34 & 97.46 & 272.44 & 563.04 & \textcolor{red}{$\shortdownarrow$1.54}\\
    \mymodel & \textbf{93.3} & \textbf{99.6} & \textbf{100} & \textbf{292.9} & \textbf{81.36} & \textbf{95.94} & \textbf{98.02} & \textbf{275.32} & \textbf{568.22} & \textcolor{blue}{$\shortuparrow$3.64}\\
    \hline
\end{tabular}
\end{table}

It was clearly that our \mymodel\ obtained the highest metrics at all recall values compared to others. \mymodel\ achieved a slightly better R@5 and R@10 in Image-to-Text (I2T) and Text-to-Image (T2I) subtasks than ALBEF. However, the gap became more significant at R@1. We improved the R@1 of I2T by $0.7\%$ ($92.96 \to 93.3$) and the R@1 of T2I by $1.6\%$ ($79.76 \to 81.36$). In total, our RSum was $3.64\%$ higher than that of ALBEF ($564.58 \to 568.22$).

The experiment also showed that LightningDOT, which encoded images using Faster-RCNN, was much behind ALBEF when its total RSum was lower than that of ALBEF by approximately $30\%$. The reason might be that the object detector was not as powerful as the ViT network and LightningDOT was pretrained on 4M images compared to 14M images used to train ALBEF. Although also using object detectors as the backbone but applying a cross-modality network, UNITER and VILLA surpassed LightningDOT by a large margin at $15\%$. It proved that this intensive architecture made the large impact on the multimodal retrieval.

Regarding our 2 baselines B1 and B2, both of them were failed to get better results than the input model ALBEF. Model B1, with the simple strategy of taking the average ranking results and having no learnable parameters, performed worse than model B2 which used a trainable linear layer to fuse the pretrained features. Nevertheless, the RSum of B2 was lower than \mymodel\ by $5.18\%$. It showed the advantages of using graph structure to fuse the information between models to obtain the better result.

\subsection{Ablation Study}
To show the stable performance of \mymodel, we used it to combine 2 other different pretrained models, including BLIP \cite{li2022blip} and CLIP \cite{radford2021clip}.
While CLIP is well-known for its application in many retrieval challenges \cite{luo2021clip4clip,tran2022mysceal,dzabraev2021mdmmt,sun2021lightningdot}, BLIP is the enhanced version of ALBEF with the bootstrapping technique in the training process.
We used the same configuration as described in \ref{writing:implement} to train and evaluate \mymodel\ in Flickr30k and MSCOCO datasets. 
We used the pretrained BLIP and CLIP from LAVIS library \cite{li2022lavis}. It was noted that the CLIP we used in this experiment was the zero-shot model, since the fine-tuned CLIP for these datasets is not available yet.

\begin{table}[ht!]
  \centering
  \caption{Performance of models on the test set in Flickr30k and MSCOCO datasets. The column $\vartriangle$R showed the difference compared to BLIP in that dataset.}
  \label{tab:ablation}
  \begin{tabular}{c|c|c c c c | c c c c|c|c}
    \hline
    \multirow{2}{*}{\textbf{Dataset}} & \multirow{2}{*}{\textbf{Methods}} & \multicolumn{4}{c}{\textbf{Image-to-Text}} & \multicolumn{4}{c|}{\textbf{Text-to-Image}} & 
    \textbf{Total} & 
    \multirow{2}{*}{$\vartriangle$\textbf{R}} \\
    \cline{3-11}
    & & R@1 & R@5 & R@10 & RSum & R@1 & R@5 & R@10 & RSum & RSum & \\
    \hline
    \multirow{3}{*}{Flickr30k} & BLIP & 94.3 & 99.5 & 99.9 & 293.7 & 83.54 & 96.66 & 98.32 & 278.52 & 572.22 & 0 \\
    & CLIP & 88 & 98.7 & 99.4 & 286.1 & 68.7 & 90.6 & 95.2 & 254.5 & 540.6 & \textcolor{red}{$\shortdownarrow$31.62}\\
    & \mymodel & \textbf{95.2} & \textbf{99.7} & \textbf{100} & \textbf{294.9} & \textbf{85.3} & \textbf{97.24} & \textbf{98.72} & \textbf{281.26} & \textbf{576.16} & \textcolor{blue}{$\shortuparrow$3.94} \\
    \hline
    \multirow{3}{*}{MSCOCO} & BLIP & \textbf{75.76} & \textbf{93.8} & \textbf{96.62} & \textbf{266.18} & 57.32 & 81.84 & 88.92 & 228.08 & 494.26 & 0 \\
    & CLIP & 57.84 & 81.22 & 87.78 & 226.84 & 37.02 & 61.66 & 71.5 & 170.18 & 397.02 & \textcolor{red}{$\shortdownarrow$97.24}\\
    & \mymodel & 75.36 & 92.98 & 96.44 & 264.78 & \textbf{58.46} & \textbf{82.85} & \textbf{89.66} & \textbf{230.97} & \textbf{495.75} & \textcolor{blue}{$\shortuparrow$1.49} \\
    \hline
\end{tabular}
\end{table}

Table \ref{tab:ablation} showed the comparison between \mymodel\ and the input models. CLIP performed worst on both Flickr30k and MSCOCO with huge differences compared to BLIP and \mymodel\ because CLIP was not fine-tuned for these datasets. Regarding Flickr30k dataset, \mymodel\ managed to improve the RSum by more than $3.9\%$ compared to that of BLIP. Additionally, \mymodel\ obtained the highest scores in all metrics for both subtasks. Our proposed framework also increased the RSum of BLIP by $1.49\%$ in MSCOCO dataset. However, BLIP performed slightly better \mymodel\ in the I2T subtask while \mymodel\ achieved higher performance in the T2I subtask.

\section{Conclusion}
In this research, we proposed a simple graph-based framework, called \mymodel, to combine 2 pretrained models to address the image-text retrieval problem. 
We created a graph structure to fuse the extracted features obtained from the pretrained models, followed by the GATv2 network to update them. Our proposed \mymodel\ only contained roughly 10M learnable parameters, helping it become easy to train using only 1 GPUs.
Our experiments showed the promisingness of the proposed method. Compared to input models, we managed to increase total recall by more than $3.6\%$.
Additionally, we implemented other 2 simple baselines to show the advantage of using the graph structures.
This result helped us resolve 2 questions: (1) increase the performance of SOTA models in image-text retrieval task and (2) not requiring many GPUs to train on any large-scale external dataset. It has opened the possibility of applying \mymodel\ in industry where many small and medium start-ups do not possess many GPUs.

Although we achieved the better result compared to the baselines, there are still rooms to improve the performance of \mymodel.
Firstly, it can be extended not only by 2 pretrained models as proposed in this research, but can be used with more than that number.
Secondly, the use of different graph neural networks, such as the graph transformer \cite{shi2020masked}, can be investigated in future work. 
Third, the edge feature in the graph is also considered. Currently, \mymodel\ did not implement the edge feature in our experiment, but they can be learnable parameters in graph neural networks.
Last but not least, pretraining \mymodel\ on a large-scale external dataset as other SOTA have done might enhance its performance.

\section{Acknowledgement}
This publication has emanated from research supported in party by research grants from
Science Foundation Ireland under grant numbers SFI/12/RC/2289, SFI/13/RC/2106, and
18/CRT/6223.

\bibliographystyle{splncs04}
\bibliography{ref}

%
%
%
%




\end{document}